\documentclass[5p,times]{elsarticle}

\makeatletter
\def\ps@pprintTitle{%
  \let\@oddhead\@empty
  \let\@evenhead\@empty
  \let\@oddfoot\@empty
  \let\@evenfoot\@oddfoot
}
\makeatother

\usepackage{graphicx}
\usepackage{color}
\usepackage{array}
\usepackage{float}
\usepackage{url}
\usepackage{enumerate}
\usepackage{amsfonts}
\usepackage{enumitem}
\usepackage{subcaption}
\usepackage{amsmath}

\newcommand{\etal}{\mbox{\emph{et al.\ }}}

\newcommand{\eg}{\mbox{\emph{e.g.\ }}}

\newcommand{\comment}[1]{}
\newcommand{\T}{{\scriptscriptstyle \top}}

\def\RR{\mathbb{R}}

\usepackage{caption,subcaption}
\usepackage{subcaption}

\usepackage{hyperref}
\begin{document}

\begin{frontmatter}

\title{Weakly and Semi Supervised Detection in Medical Imaging via Deep Dual Branch Net}

\author{Ran Bakalo$^{1,2}$}
 \author{Jacob Goldberger$^3$\corref{mycorrespondingauthor}}
\cortext[mycorrespondingauthor]{Corresponding author}
\ead{jacob.goldberger@biu.ac.il}
 \author{Rami Ben-Ari$^2$\corref{}}

\address{$^1$Department of Computer Science, University of Haifa, Israel}
\address{$^2$IBM Research, Haifa, Israel}
\address{$^3$Faculty of Engineering, Bar-Ilan University, Ramat-Gan 52900, Israel}
\date{}

\begin{abstract}
This study presents  a novel deep learning architecture for multi-class classification and localization of abnormalities in medical imaging illustrated through experiments on mammograms.  The proposed network combines two learning branches. One branch is for region classification with a newly added normal-region class. Second branch is region detection branch for ranking regions relative to one another. Our method enables detection of abnormalities at full mammogram resolution for both weakly and semi-supervised settings. A novel objective function allows for the incorporation of local annotations into the model. 
We present the impact of our schemes on several performance measures for classification and localization, to evaluate the cost effectiveness of the lesion annotation effort. Our evaluation was primarily conducted over a large multi-center mammography dataset of $\sim$3,000 mammograms with various findings. The results for weakly supervised learning showed significant improvement  compared to previous approaches.  
We show that the time consuming local annotations involved in supervised learning can be addressed by a weakly supervised method that can leverage a subset of locally annotated data.
Weakly and semi-supervised methods coupled with detection can produce a cost effective and explainable model to be adopted by radiologists in the field.
\end{abstract}


\end{frontmatter}

% \linenumbers

\section{Introduction}
The most common type of cancer and the second leading cause of death in women is breast cancer \cite{jemal2011global}. Nearly 40 million mammography exams are performed on a yearly basis in the
US alone. Screening mammograms (MG) are the first line of imaging for the early detection of breast cancer. These raise the survival rate, but place a massive workload on radiologists. Although mammography provides a high resolution image, its analysis remains challenging because of tissue overlaps, the high variability between individual breast patterns, subtle malignant findings (often less than 0.1\% of the image area) and the high similarity between benign and malignant lesions. 
Suspicious lesions are often difficult to detect and classify, even by expert radiologists. Lesions can be relatively small with respect to the whole image and occluded in the parenchymal tissues. 

A broad range of traditional machine learning classifiers have been developed for automatic diagnosis of specific findings such as masses and calcifications, and ultimately breast cancer \cite{de2015classification,jen2015automatic}. Ultimately, diagnosis in mammograms is often dictated by the type of lesion found. 

Our goal is building an automatic system that
can jointly detect the lesion location  (if it exists) and analyze  the findings. This goal can be achieved by training a detector from local (often referred as instance) annotations \cite{Lotter_MICCAI17,D_Ribli2018}, and then classifying the image according to the most severe finding in the image. However, in this type of supervised setting, training requires bounding-box annotations for every single abnormality. This setting is tedious, costly and impractical for large data sets. This problem is exacerbated in mammograms that can contain tens or hundreds of micro-calcifications spread throughout the breast. Having manual annotations further increases the likelihood of inconsistency in labeling due to a lack of consensus between radiologists \cite{QuaMaDi2007} caused by ambiguous lesion boundaries.  This problem is often resolved by having multiple annotators \cite{DBLP:conf/aaai/GuanGDH18} that further escalates the workload.

\begin{figure*}[ht!]
\centering
\subcaptionbox{Normal}
{\includegraphics[width=2.4 cm, height=3.36 cm]{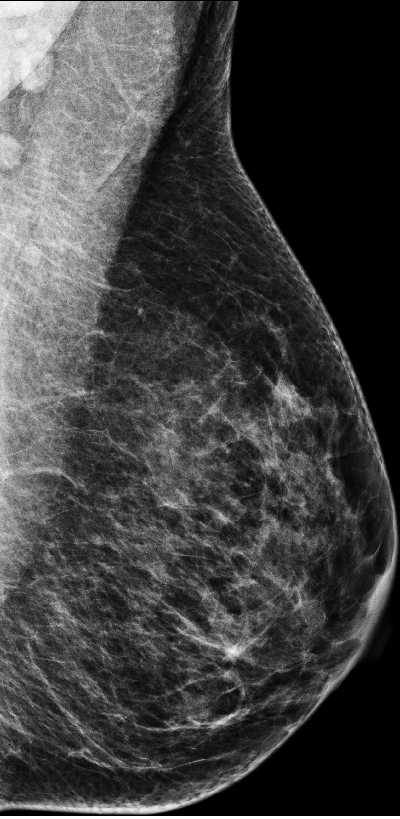}}
\subcaptionbox{Benign}
{\includegraphics[width=2.4 cm, height=3.36 cm]{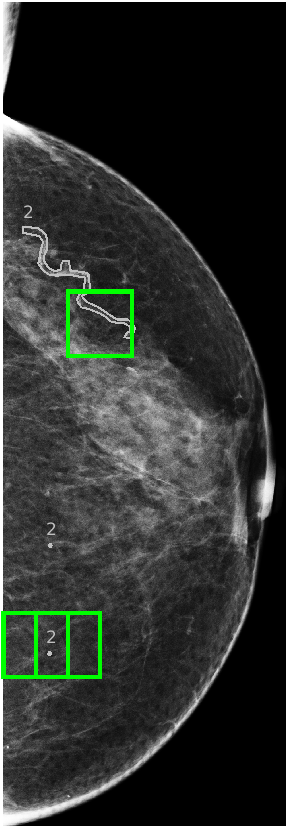}}
\subcaptionbox{Malignant}
{\includegraphics[width=2.4 cm, height=3.36 cm]{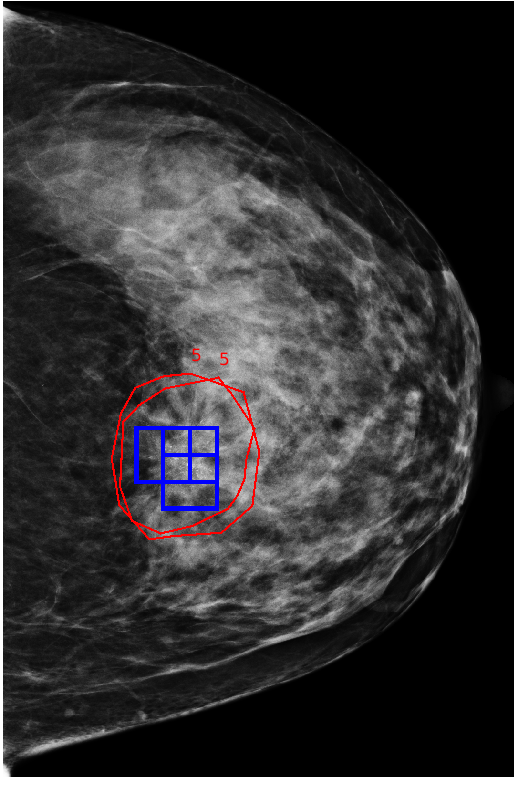}}
\subcaptionbox{Malignant + Benign}
{\includegraphics[width=2.4 cm, height=3.36 cm]{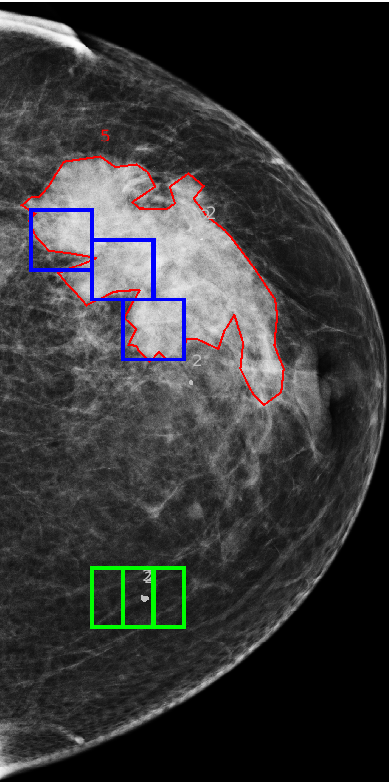}}
\caption{Illustration of the classification and localization task. The input is the corresponding mammogram and the output is the class (Normal/Benign/Malignant) and the finding localizations. The radiologist's local annotations are shown as contours (red for malignant and gray for benign). The class and the model's local predictions are shown as colored bounding boxes (Normal mammograms do not have a bounding box, Benign regions are in green and Malignant in blue). The image Malignant+Benign is a case with additional benign finding.}
\label{fig:Illustration}
\end{figure*}

In the weakly supervised paradigm, only global image-level tags are provided to train a classifier. Global image labels are easily available from retrospective clinical records often without the need for further clinician intervention. 
Weak supervision, however,  provides no local  information on the lesion location. 
 In an era of growing demand for XAI (explainable AI), localization can shed light on the model reasoning for the image classification, and help foster trust among practitioners in the field.
 Hence, weakly supervised methods which also localize abnormalities provide high value especially in scenarios where the source of discrimination between the classes is a-priori unknown.

In this study we address the acute problem of annotation and suggest a new network that can be trained on weakly labeled data and is capable of localizing the lesions at test time (perform detection), in full resolution. 
 Our network architecture is composed of two branches (streams), one for classification and the other for detection. In the classification branch regions are classified to abnormality classes (\eg benign or malignant) and a newly added {\it normal-region} class representing healthy tissues. In the detection branch the scores of all regions are ranked relative to one another, for each abnormality class (resulting in a distribution over regions per abnormality class). The classification branch classifies each region, whereas the detection branch selects which regions are more likely to contain a finding. The image class probability is then obtained by aggregation of the detection and classification probabilities for all regions in the image. The final abnormality probability is then increased when a suspicious finding is contained in one of the regions, similar to a radiologist's inspection work flow.

 The main contributions of this work are as follows: 1. A dual branch deep learning architecture for joint image classification and region detection via region classification branch with a newly added normal-region class and, in parallel, region ranking branch.
 2. A weakly-supervised learning method to train the proposed network. 
 3. A semi-supervised learning method to train the proposed network. Our method enables joint learning using weakly supervised data and additional fully supervised data with a novel region-level objective function on the branches' region-level probabilities.

Semi-supervised datasets combine globally labeled data with a small amount of data with explicit local annotations in addition to the global labels.
In this work, we explore the problem of training the described network in weakly supervised and semi-supervised setups. The results of the proposed system are  illustrated in Fig. \ref{fig:Illustration}.

We validate our method on a large FFDM dataset of nearly 3,000 mammograms as well as the public INBreast dataset \cite{moreira2012inbreast}. Direct comparison of our method to previous works \cite{choukroun2017mammogram,bilen2016weakly} and an ablation study shows that our model outperforms others in classification and, in particular, in detection.

A preliminary version of this work (with only weakly supervised setting) has been reported \cite{dualBranchISBI}. Our study include additional results analysis, ablation study and addition of a semi-supervised learning method.
\section{Related Work}

Deep learning methods promise a breakthrough on assisting breast radiologists for early cancer detection in mammograms. However, the bottleneck for supervised methods in Big Data is the annotation workload which often requires expert clinicians/radiologists to delineate numerous benign as well as malignant findings in mammograms. Weakly supervised and semi-supervised methods are considered an affordable compromise to this tangle.

\subsection{Weakly supervised detection}
Weakly supervised detection methods in deep learning have attracted growing interest with the publish of the paper ``Is localization for free?" \cite{Oquab2015} that addressed the tedious task of local annotations in images \cite{bilen2016weakly,jiang2017optimizing}. Recent studies and challenges in mammography that have vast datasets (of over 0.5 million mammograms) have opted for weakly labeled data \cite{DreamChallenge, Geras2017}. 

In general, there are two main approaches to weakly supervised learning, known as image and region based. In image based methods based on CNN \cite{hwang2016self,old_DBLP:conf/miccai/ZhuLVX17}, the input to the model is the whole image. Region inference is then obtained from feature maps after pooling at the final CNN layer (often generating a heat map). In region based methods \eg \cite{choukroun2017mammogram,DBLP:journals/tmi/YanZPLSZMZ16}, the image is first decomposed into regions. The convolutional layers then process each region separately. Subsequent layers then classify the regions and aggregate results to a global class level.

\textbf{Image based.}  Zhu \etal \cite{old_DBLP:conf/miccai/ZhuLVX17} proposed an image based method for mammogram classification based on Multi-Instance Learning (MIL) that classifies large tiles of the image by max-pooling over feature maps, with sparsity soft constraints. However, when using down-sampled images, their method yielded detection maps with a low resolution of just 6$\times$6 pixels, which curtails practical use considerably. 
Hwang \etal \cite{hwang2016self} also took an image-based approach using a CNN with two whole image classification branches that shared convolution layers. One branch used fully connected layers, and the second branch used $1\times1$ convolution layers, resulting in one map per class, and then a global max pooling on each map. Their method yielded a low AUROC of 0.65 over 332 MIAS mammograms. 
Both  of these image-level studies \cite{old_DBLP:conf/miccai/ZhuLVX17,hwang2016self} addressed a binary classification task with a small test set of 410 full-field digital mammography (FFDM) mammograms \cite{old_DBLP:conf/miccai/ZhuLVX17} or using non-FFDM (digitally scanned) images of the MIAS cohort instead.

\textbf{Region based.} Yan \etal \cite{DBLP:journals/tmi/YanZPLSZMZ16} proposed a region-based method for a different use-case of discriminating between local anatomies in CT scans, using MIL in a DNN setting. Choukroun \etal \cite{choukroun2017mammogram} recently implemented a region-based approach with the MIL paradigm to classify the entire mammogram according to the max-probability region, thus also providing detection in {\it full resolution}.

These methods \cite{choukroun2017mammogram,old_DBLP:conf/miccai/ZhuLVX17,DBLP:journals/tmi/YanZPLSZMZ16} apply an implicit detection regime via a max-pooling operation on regions or region classification probabilities. 

\textbf{Dual branch architecture.} Recent studies on natural images, suggest that applying explicit {\it data-driven} detection in parallel to classification yields improved performance \cite{bilen2016weakly}. In this study we follow a region-based dual branch approach, but differ and generalize the existing method \cite{bilen2016weakly} in two main ways: 1) We don't use any unsupervised region proposal in our scheme as it is commonly unavailable in medical imaging. 2) We adapt the architecture in \cite{bilen2016weakly} and extend the {\it region} classification stream to include an additional \textit{normal-region} class but without any  detection counterpart. This makes it possible to handle images without any findings (objects). In addition, it reduces the false positives resulting from normal regions in detection. Also, this enable to use the network in a semi-supervised detection setting with joint learning from weakly and fully supervised data. The extension for handling radiology images assessed as normal is equivalent to images without any objects in natural images. The addition of a normal-region class changes the probability distribution for the regions, and allows improved classification of these specific and prevalent normal cases in many medical use cases such as screening mammography. Similar to \cite{jiang2017optimizing}, we further connect the branches by adding information from the classification branch to guide the detection branch to the most relevant regions. 

Our model is capable of multi-class classification and detection that provides localization of the abnormalities in full resolution. We compare our method to the one described in \cite{choukroun2017mammogram} and an approach based on \cite{bilen2016weakly}. We report improved performance in both classification and detection. 

\subsection{Semi supervised detection}
Semi supervised detection methods involve the fusion of weak labels with a subset of data having local annotations, namely {\it fully labeled} (also known as strongly labeled) data. There are two main approaches to semi-supervised detection setting. The first approach is two-stage training with a stage for fully labeled data and a stage for weakly labeled data. The second approach involves joint training from weakly and fully supervised data.

\textbf{Two-stage approach.} A large data set of fully labeled data with lesion annotations is used to train a region based classifier. Then, at a subsequent stage the model is modified for whole image input (usually decomposed into regions) and fine-tuned on the weakly labeled data to create a weakly labeled classifier \cite{D_Ribli2018,LiShenArxiv2018,Lotter_MICCAI17,shen2019deep}. However, these methods rely strongly on local annotations and need a sufficiently large fully labeled dataset to initialize the model. They are unable to train solely on weakly labeled mammograms and often lack detection capability (except Ribli \etal \cite{D_Ribli2018} that uses detection based on instance labels). 

Wu \etal \cite{DBLP:journals/corr/abs-1903-08297} used patch-level classifier for producing heat maps as additional input channels to a multi-view breast-level classifier.

\textbf{Joint training from weakly and fully supervised data.} A single model is trained jointly using weakly and fully supervised data by combining a weakly supervised objective function with a fully supervised objective function.

Yan \etal \cite{DBLP:journals/corr/YanLPLZ17} proposed a method for weakly supervised training of Fast RCNN \cite{DBLP:conf/iccv/Girshick15} via Expectation Maximization (EM). Focusing on the detection problem, they treated instance-level (region level) labels as missing data for weakly annotated images. Their method alternated between two steps: 1) E-step: estimating a probability distribution over all possible latent locations in weakly supervised images, and 2) M-step: training Fast RCNN using estimated locations from the last E-step. They proposed a semi-supervised learning method by adding a standard fully supervised objective function to the fully supervised images which are then used to train the Fast RCNN network in the M-step in addition to the weakly supervised objective function. Their method was applied on non-medical (natural) images, and in practice, the quality of the solution depended heavily on initialization by another method (\cite{bilen2016weakly}, which we compare our method with). Furthermore, their approach required thousands of Fast RCNN training iterations at each M-step, which is computationally expensive, particularly for large images such as mammograms.

Cinbis \etal \cite{DBLP:journals/pami/CinbisVS17} suggested a MIL approach for weakly supervised detection in natural images. They suggested extending their method to a semi-supervised setting by replacing the top region selection, obtained from MIL, with the ground-truth regions when training from fully-supervised images.

In the medical domain, an approach based on Faster RCNN \cite{DBLP:journals/pami/RenHG017} was taken by Shin \etal \cite{DBLP:journals/corr/abs-1710-03778}, and was applied to breast Ultrasound (US) images. They also proposed semi-supervised training, but based on combination of Faster RCNN \cite{DBLP:journals/pami/RenHG017} and MIL. However, in breast US, only the field of view with suspicious masses were considered (and not calcifications or images without any abnormality). Unlike mammograms, a lesion in an US captures a relatively large area of the image. Mammography therefore appears to be a greater challenge in that there are more types of lesions with a significantly lower signature.

Li \etal \cite{DBLP:conf/cvpr/Li0HXWLF18} proposed a semi-supervised classification and detection method for chest x-ray images. In their model, the input image is processed by CNN. Then, there is max pooling or interpolation on the feature maps to get patch grid, which is then processed by a fully-convolutional recognition network, resulting in patch scores for multiple categories. Then, they have global predictions based on MIL criterion. They define the global positive probability by the complement of the joint probability of all the patches being negative, assuming patches probabilities are independent of each other. They combine a fully supervised loss function on the fully supervised images and a weakly supervised loss function on the weakly supervised images.

 We follow the joint training approach. In our approach, the local annotations are used as auxiliary data, and our model can be trained with a small fully annotated dataset, mostly relying on weak labels. Given the high cost of annotation in many medical domains, we believe that this approach can provide a competitive edge.

\section{A dual branch weakly supervised detection methodology for mammograms} 
\label{WeaklyMethod}
In this section we propose a deep network architecture that classifies mammogram regions into three different classes: normal tissue, benign, and malignant findings using labels at the image level (also known as {\it weak} labels).

\begin{figure*}[!ht]
\centering
\includegraphics[scale=.35]{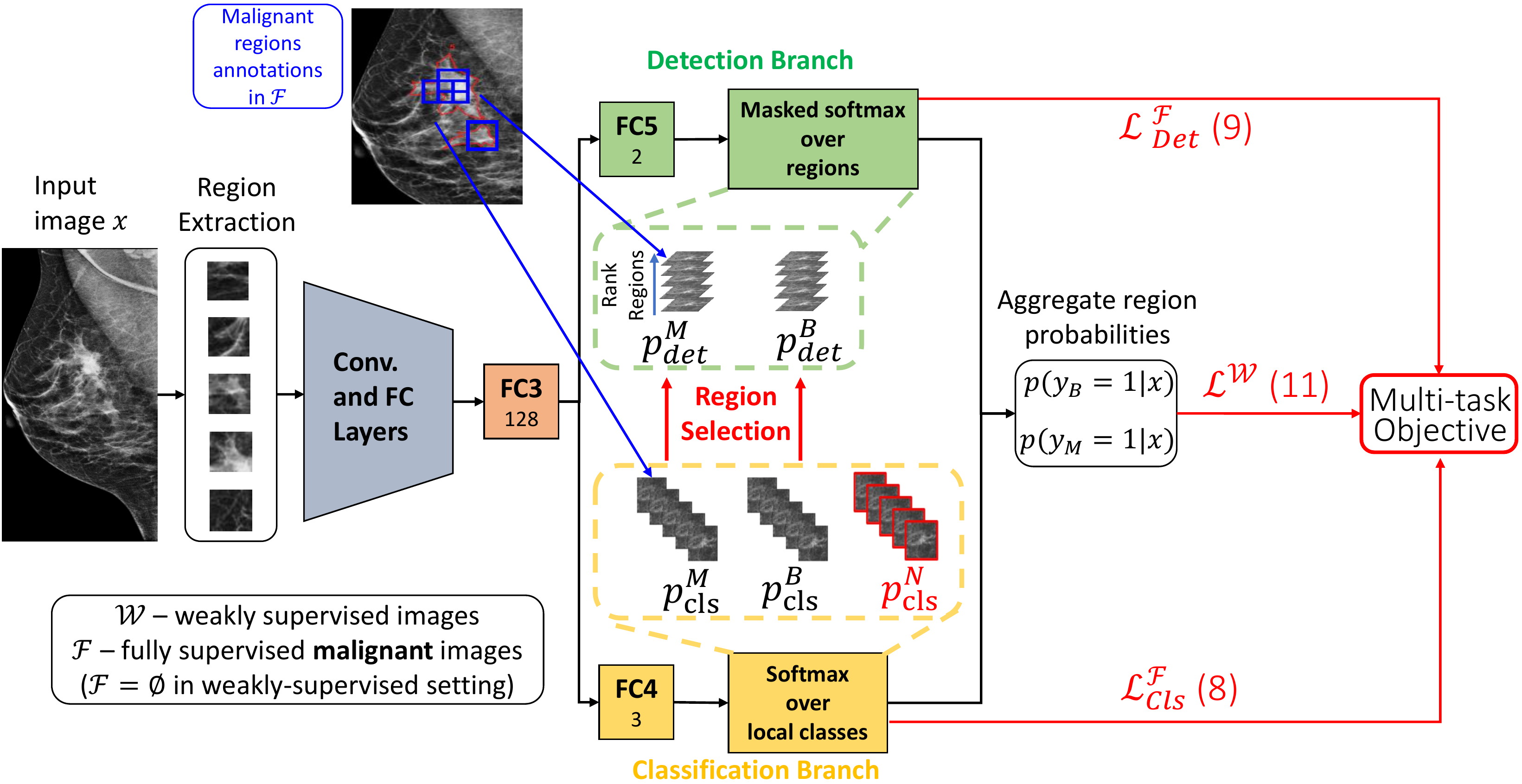}
\caption{Architecture overview.  The novel elements are in red. Our new scheme has an additional class in the classification branch ($p_{cls}^N$ ) with no associated detection and a region selection model connecting the branches. FC blocks show fully-connected layers with the corresponding output size. There are two FC layers in the ``Conv. and FC layers" block from the pre-trained VGG. Layer FC3 is shared between branches, FC4 is for the classification branch, and FC5 is for the detection (ranking) branch. $\mathcal{L}^{\mathcal{F}}_{Cls}$, $\mathcal{L}^{\mathcal{F}}_{Det}$ and $\mathcal{L}^{\mathcal{W}}$ corresponds to Eq. \ref{eq:l_f_cls}, \ref{eq:l_f_det} and \ref{eq:l_w} respectively.}
\label{fig:WeakArchitecture}
\end{figure*}

We first decompose the image into regions that are fed into the network. The network has two branches: a {\it classification}  branch that computes local probabilities of malignant, benign and normal for each region, and a {\it detection} branch that ranks regions relative to one another for the malignant class and, independently, for the benign class. The branches are then combined at a subsequent layer to obtain an image-level decision for the presence of malignant and/or benign findings. The proposed weakly supervised network architecture is depicted in Fig. \ref{fig:WeakArchitecture}, and the algorithm is summarized in Table \ref{alg_tbl}.

\textbf{Region extraction.}
 Given a mammography image, we first perform pre-processing to compute feature representations for regions within the breast.
 To this end, we used a sliding window of $224 \times 224$ overlapping regions (with a $112 \times 112$ stride) within the breast region excluding the axilla (using a method similar to \cite{Chunxiao2015_Axilla}).
 
 Due to the relatively small training dataset, we employ a two-stage deep neural network architecture. In the first stage, we apply a transfer learning approach by using the pre-trained VGG128 network \cite{Chatfield14}, trained on the ImageNet dataset \cite{deng2009imagenet}. In our model, we extract CNN codes from the last hidden layer as 128D feature vectors per region.
Then, we process each region separately by a fully connected (FC) layer. Formally, an image $x$, is first decomposed into $m$ regions denoted by  $r_1,...,r_m$ such that  $\phi(r_i) \in \RR^{128}$ is the feature vector representation of the $i$-th region.

\textbf{Classification branch.} We first compute a local decision for each region separately. Each region is classified, in this study, as  normal (N), benign (B) or malignant (M) using a softmax layer:
\begin{equation} \label{class_br}
\begin{split}
   &p_{\mbox{cls}}(c|r_i) =  \frac{ \exp (w_c^{\T} \phi(r_i))} { \sum_{d\in\{N,B,M\}}  \exp (w_d^{\T} \phi(r_i))}, \\ 
 &c\in\{N,B,M\}, \quad i=1,\ldots,m
\end{split}
\end{equation}
such that $w_N, w_B$ and $w_M$ are the parameters of the classifier. Note that the same classification parameters are used for all the regions in the image.

\textbf{Detection branch.} In parallel, we compute the relevance of each region for the global image-level decision. We perform a separate detection process for each type of abnormality - one for malignant regions and one for benign regions. The normal class has different characteristics. These regions are prevalent in all types of mammograms, similar to the ``background" in natural images. Therefore, the normal class is not associated with a detection scheme (see Fig. \ref{fig:WeakArchitecture}). This is a novel extension to previous modeling in \cite{bilen2016weakly}. In \cite{bilen2016weakly} the image-level class set and the region-level class set are the same and are used in both branches. The detection result is a distribution for the malignant class and a distribution for the benign class. Each such distribution is over all the regions in the image implemented by a softmax operation. Formally, let $z_c$ be a hidden random variable representing the localization of class $c$  findings in the image. Then, given an image $x$, the probability of $z_c = i$ in the $c$ distribution is:
\begin{equation}
p^c_{\mbox{det}}(i|x) = \frac{ \exp(u_c^{\T} \phi(r_i))}{\sum_{j=1}^m \exp ( u_c^{\T} \phi(r_j)) }, \quad
c\in\{B,M\}, \quad i=1,\ldots,m
\end{equation}
such that $u_B$ and $u_M$ are  the parameter-sets of the benign and malignant detectors, respectively. Note that $p^c_{\mbox{det}}(i|x)$ is equivalent to the ranking of the $i$-th region in image $x$ relative to the other regions in $x$ for class $c$.

\textbf{Image level decision.} Given the region-level classification results and the region detection distribution, we can now evaluate the image-level classification.
Let $(y_M$, $y_B)$ be a binary tuple indicator whether an image contains a malignant and/or benign finding, respectively. Note that this type of tuple labeling allows for tagging images of class $N$ by $(0,0)$ and those with both $M$ and $B$ findings by $(1,1)$.
The posterior distributions of $y_M$ and $y_B$ given  mammogram image $x$ are obtained as a weighted average of the local (i.e. region-level) decisions:
\begin{equation}
p(y_c=1|x) = \sum_{i=1}^m p^c_{\mbox{det}} (i|x) p_{\mbox{cls}}(c|r_i), \quad c\in\{B,M\}.
\label{Eq:GlobalScores}
\end{equation}
\textbf{Comparison to previous dual-branch approach.} Since in many medical applications such as mammography, the most prevalent cases are normal without any findings, we extended the method in \cite{bilen2016weakly} by adding a normal-region ($N$) class to the classification branch. Note that in our new scheme the normal class is only added to the classification branch and not to the detection branch or to the image-level class set (see Fig. \ref{fig:WeakArchitecture}). This is a novel generalization to previous modeling in Bilen \etal \cite{bilen2016weakly}. In oppose to Bilen \etal \cite{bilen2016weakly}, by allowing classification of \textit{regions} to normal, we can handle ``clean" images without any findings. Normal images in our model are then discriminated by having a low probability for both $M$ and $B$ findings. The probability for an image to be normal can then be obtained via the joint probability $p(y_M=0,y_B=0|x)$.

This extension is also important for reducing the false positives in detection (localization) resulting from normal regions, as shown in Sec. \ref{sec:exp}, since normal regions gain high probability for local class $N$ and low probabilities for $M$ and $B$ (instead of expected uniform probabilities over $M$ and $B$ when the $N$ class is not used \cite{bilen2016weakly}).

In addition, this modified architecture enables the use of a fully supervised loss on the classification branch for the extension to a semi-supervised detection setting in Sec. \ref{sec:semi}

{\bf Region selection.} So far the detection branch's decision has solely been based on the features that were extracted from the image regions. It makes sense to use the classification
decision results to guide the detection process. For example, if a region is clearly classified as malignant, it is likely that the malignancy
detection will favor this region. Since the classification branch includes an additional class for normal regions, the suspicious regions in the B and M classes can be used to guide the detection branch and create a soft alignment between the branches.  We formalize this intuition by a  region selection step. Now, let
$p_{\mbox{cls}}(M|r_1),..., p_{\mbox{cls}}(M|r_m)$ be the region  probabilities of being  classified as malignant.
In the malignant detection  process, we only consider the  $k$ regions with the highest probability of being  classified as malignant and only apply  the softmax operation  on these selected regions.  Let $h_M(i)$ be a binary value indicating whether region $i$ has been
selected for the malignancy detection process. We can apply the same selection criterion to the benign detector. Thus, each detector's ranking is conducted solely on the relevant regions according to the classification  branch.
In the  modified detection branch we replace the softmax over regions by a masked softmax:
\begin{equation}
\begin{split}
&p^c_{\mbox{mask-det}}(i|x) = \frac{ h_c(i) \exp(u_c^{\T} \phi(r_i))}{\sum_{j=1}^m h_c(j) \exp ( u_c^{\T} \phi(r_j)) }, \\
& c\in\{B,M\},  \quad i=1,\ldots,m.
\label{maskdet}
\end{split}
\end{equation}
This paradigm guides the M detector to focus on the most probable malignant regions in malignant mammograms.  However, if the image is normal or contains a benign finding, the model will concentrate on regions that were most probably and erroneously classified as malignant (hard negatives). This process, which is applied similarly to the benign class is equivalent to {\it hard negative mining}.
In the experimental section we compare network architectures with and without masked detectors and show that applying region selection yields superior performance.

\begin{table}[t!!!]
\normalsize
\caption{The proposed weakly supervised detection method.}
\label{alg_tbl}
 \centerline{\fbox{\parbox{3.4in}{ \vskip0.2cm
{\bf Input data}:  an image $x$ decomposed into $m$ regions $r_1,...,r_m$ each represented by 128 features computed by a VGG network. \\ \\
{\bf Algorithm}:
\begin{itemize}
\item  Region level classification into  classes $C=\{N,B,M\}$:
$$ p_{\mbox{cls}}(c|r_i) =  \frac{ \exp (w_c^{\T} \phi(r_i))} { \sum_{d\in C}  \exp (w_d^{\T} \phi(r_i))},\hspace{0.6cm}
 c\in C $$
\item For $c\in\{B,M\}$:
\begin{itemize}
\item Mask Computation: \\
  $\,\, h_c(i)$ is a binary value indicating whether region $i$ is one of the  $k$ regions with the highest probability of being  classified as $c$.
\item Selected regions ranking:
$$
p^c_{\mbox{mask-det}}(i|x) = \frac{ h_c(i) \exp(u_c^{\T} \phi(r_i))}{\sum_{j=1}^m h_c(j) \exp ( u_c^{\T} \phi(r_j)) }
$$
\item Image level decision:
$$
p(y_c=1|x) = \sum_{i=1}^m p^c_{\mbox{mask-det}} (i|x) p_{\mbox{cls}}(c|r_i)
$$
\end{itemize}
\end{itemize}
  }}}
\end{table}

\textbf{Training.} Assume we are given a set of $n$ weakly labeled mammography images $\{x(1),...,x(n)\}$. Each image $x(t)$ consists of regions $\{r_1(t),...,r_m(t)\}$ and is associated with a binary tuple label $(y_M(t),y_B(t))$ that indicates whether the image contains at least one malignant and/or one benign finding respectively. A normal case will have a $(0,0)$ label whereas a mammogram with both M and B finding will be labeled $(1,1)$. The network provides soft decisions for each image $x(t)$   regarding the values of $y_M(t)$ and $y_B(t)$.
The objective function that we  maximize in the network training step is the following likelihood function:
\begin{equation} \label{eq:MC_loss}
L(\theta) =  \sum_{c \in \{M,B\}} \sum _{t=1}^n \log p( y_c(t) | x(t); \theta)
\end{equation}
such that $\theta$ is the parameter-set of the model (which includes the fully connected layer $\phi$ and the parameters $w$ and $u$) and the probability  $p( y_c(t) | x(t); \theta)$ is defined in Eq. (\ref{Eq:GlobalScores}).

\section{Semi supervised detection methodology} \label{sec:semi}
\subsection{Approach Overview}
In this section, we extend our weakly supervised setting to a novel semi-supervised approach. 
In a semi-supervised setting, we assume that part of the weakly labeled data has been subjected to local annotations, thus generating a subset of {\it fully-labeled} data. This local annotation can take the form of contours around lesions or simply bounding boxes. We demonstrate our model on M vs. B $\cup$ N. To reduce the annotation workload, let us assume that the malignant class has a fully-labeled subset in which only the malignant findings are locally annotated (note that malignant images can still include benign findings).

We make use of different ratios of local annotations in the malignant class (25\%-100\%) to present the impact of these annotations on performance. Due to the rarity of malignant findings with respect to benign ones, the annotated set only captures 2.5\%-10\% of all the lesions in the cohort, therefore demanding a low workload for annotation.

Our dual-branch approach differs from previous approaches \cite{DBLP:journals/corr/YanLPLZ17,DBLP:journals/corr/abs-1710-03778,DBLP:conf/cvpr/Li0HXWLF18} in architecture and objective function. Our semi-supervised method is different from previous methods by having a region ranking branch in the architecture. In addition, previous methods \cite{DBLP:journals/corr/YanLPLZ17,DBLP:journals/corr/abs-1710-03778,DBLP:conf/cvpr/Li0HXWLF18} added a fully-supervised objective function on the region classification in fully-supervised images subset. In our method, we add a fully-supervised objective function on the region classification, and, in addition, we add a fully-supervised objective function on the detection (ranking) branch's region probabilities of the fully-supervised images subset.
\subsection{Semi-Supervised Detection Objective Function}
Although local annotations on a large scale are commonly out of reach \cite{Geras2017}, in this section we examine the effect of engaging with a small set of locally annotated data combined with a large set of weakly labeled data.
We assume that the training set contains two distinct sets, one with weakly and one with fully labeled images. We denote  $\mathcal{W}$ as the set of indices of the weakly-labeled images (these can be malignant, benign or normal) and $\mathcal{F}$ as the set of indices of the {\it fully-labeled} images; namely,  mammograms where lesions have been locally annotated. For each fully labeled image, $x$, we are given a set $\mathcal{M}_x$ of malignant regions. We next describe how we transform the pixel-level information (i.e. contour annotations) into the region-level labels based on the intersection  between our extracted regions and the malignant lesion. To this end, we define a soft version of Intersection over Union ($IoU$) called the {\it Intersection over Minimum} ($IoM$). This measure computes the ratio between the area of the intersection with respect to the minimum  size between the $i$-th region $r_i$ and the lesion area:
\begin{equation}
IoM(i,c) = \frac{|r_i \cap c|}{\min \{|r_i|,|c|\}},
\label{eq:IOM}
\end{equation}
where $c$ is the annotated domain. In our setting the region size is fixed and the lesion scale can vary by a factor of 10. This definition therefore allows a positive region to cover a small lesion or alternatively be located within a large finding. We define the local label of a region as malignant ($M$) if the region has $IoM \geq \alpha$ with a ground-truth (GT) \textbf{malignant} finding, and define the label as either benign or normal ($BN$) if the region has an empty intersection with all the GT malignant findings. We set $\alpha=0.5$. Formally, the label of region $r_i$, denoted by  $y_i$, is defined as follows:
\begin{equation}
y_i = \left\{
                 \begin{array}{ll}
                   M \ \ \ \ \ \ \exists \ c \in \mathcal{M}_x \ \ s.t.  \ IoM(i,c) \geq \alpha\\ \\
 BN\ \ \ \ \forall \ c \in \mathcal{M}_x, \ r_i \cap c = \emptyset
                 \end{array}
               \right.
 \end{equation}
Non-malignant regions with $IOM < \alpha$ are ignored during training. In practice, we achieved better performance when ignoring those regions during training compared to labeling the regions as $BN$.

In order to engage the local annotations, we propose two separate and novel objective functions that are imposed directly on the region classification and detection probabilities. In the fully supervised objective of the classification branch, we compute the log likelihood according to the region true classes (as $M$ or $BN$) as:
\begin{equation}
\mathcal{L}^{\mathcal{F}}_{Cls}(\theta) = \sum_{t\in \mathcal{F}} \sum_i
\log p_{\mbox{cls}}(y_i(t)|r_i(t))
\label{eq:l_f_cls}
\end{equation}
where  $t$  goes over all the fully labeled images, and $i$ goes over the labeled regions in each image.
   The  probability of a region to be classified as malignant, $p_{\mbox{cls}}(M|r_i(t))$, is defined in Eq. (\ref{class_br}), and   $p_{\mbox{cls}}(BN|r_i(t))$ is the complement probability (i.e., the probability of being classified as either benign or normal).

In the fully-supervised objective of the detection branch, we want to concentrate on the malignant regions. We therefore define the detection branch objective as:
\begin{equation}
\mathcal{L}^{\mathcal{F}}_{Det}(\theta) =  \frac{1}{|\mathcal{F}|} \sum_{\substack{t \in \mathcal{F}}} \log \left( \sum_{\substack{i | y_i(t) = M}} p^M_{det} \left( i|x\left(t \right) \right) \right).
\label{eq:l_f_det}
\end{equation}
This demanding regions with high overlap over M-lesions to have high M-probability. This soft constraint alters the weakly supervised decisions toward manually labeled regions. The trained model eventually relies on discriminative power and similarity to the annotated regions as the source of malignancy when making its decisions. 

Without loss of generality, we assume the fully-supervised objective is applied on the malignant images in $\{x(t) : t \in \mathcal{F}\}$. Our final fully supervised objective is then obtained as:
\begin{equation}
\mathcal{L}^{\mathcal{F}}(\theta) = \lambda_1 \mathcal{L}^{\mathcal{F}}_{Cls}(\theta) + \mathcal{L}^{\mathcal{F}}_{Det}(\theta).
\label{eq:fullySupervisedObj}
\end{equation}
We set $\lambda_1=\beta/m_f$ where $m_f$
is the total number of regions in the train data that have a region-level label. For simplicity, we set $\beta=1$. 

The weakly supervised part, $\mathcal{L}^{\mathcal{W}}$, is defined in a similar way as in Sec. \ref{WeaklyMethod}, Eq. \ref{eq:MC_loss}. In the semi-supervised setting, this objective is defined over the weakly labeled training subset for the M class and over all the images for the B class:
\begin{equation}
\mathcal{L^{\mathcal{W}}}(\theta) =  \frac{1}{|\mathcal{W}|}\sum_{\substack{t \in \mathcal{W}}} \log p(y_M(t)|x(t);\theta)  +\frac{1}{n} \sum_{t=1}^{n} \log p(y_B(t)|x(t);\theta) 
\label{eq:l_w}
\end{equation}
In order to prevent redundancy in the training samples we avoid using the fully labeled images also as weakly labeled samples, since they were shown to degrade performance in Shin \etal \cite{DBLP:journals/corr/abs-1710-03778}.

The fusion of the weakly and fully supervised settings can now be achieved by maximizing the following multi-task objective:
\begin{equation}
\mathcal{L}(\theta) = \mathcal{L}^{\mathcal{W}}(\theta) + \lambda_2 \mathcal{L}^{\mathcal{F}}(\theta)
\end{equation}
where $\mathcal{L}^{\mathcal{W}}$  denotes the \textit{weakly} supervised part and  $\mathcal{L}^{\mathcal{F}}$ denotes the \textit{fully} supervised part.

\section{Experimental results } \label{sec:exp}

\subsection{Experiment Setup}

{\bf Dataset.} We conducted experiments on a large screening dataset, named IMG, with full field digital mammography (FFDM). The cohort was acquired from different Hologic devices and 4 different medical centers (with approximately 3K $\times$ 1.5K image size). From this proprietary dataset we excluded images containing artifacts such as metal clips, skin markers, etc., as well as large foreign bodies (pacemakers, implants, etc.). Otherwise, the images contain a wide variation in terms of anatomical differences, pathologies (including benign and malignant cases) and breast densities that corresponds to what is typically found in screening clinics. The dataset was composed of 2,967 mammograms with normal images as well as various benign and suspiciously malignant findings. In terms of the {\it global} image BI-RADS (Breast Imaging Reporting and Data System), we had 350, 2,364, 146 and 107 corresponding to BI-RADS 1,2,4 and 5 captured from 65, 693, 81 and 62 individuals respectively. Note that our BI-RADS 1 (Normal category) did not contain any suspicious findings, or confidently benign ones. Since a mammogram can contain findings with different BI-RADS categories, the global image BI-RADS was set by the most severe finding in the image (max operation), and the global patient BI-RADS was set by the max global image BI-RADS for that patient in a specific study, according to clinical guidelines.

Mammograms with global BI-RADS of 3 were excluded from our IMG dataset since these intermediate BI-RADS are commonly assigned based on other modalities (\eg ultrasound) and comparison to prior mammograms \cite{excludeBIRADS3Justification2016} which are often unavailable.
However, our data set included BI-RADS 3 findings that were not the most severe ones in the image. In terms of breast composition, 20\% were ``almost entirely fatty", 48\% had a ``scattered fibroglandular density", 27\% were ``heterogeneously dense" and 5\% were ``extremely dense". With respect to the dominant pathologies, our data set included 4525 calcifications (micro and macro) and 926 masses.

In our test scenario, we split the mammograms into the following three global labels: BI-RADS 4 \& 5 were defined as malignant (M), BI-RADS 2 were defined as benign (B) and BI-RADS 1 as normal (N). We included all types of suspiciously malignant abnormalities in the M class such as mass, calcification, architectural distortions etc. This discrimination in data classes creates a specific challenge, demanding the model to distinguish between images with very similar types of lesions, such as malignant versus benign masses or different types of micro-calcifications that are often ambiguous even for expert radiologists. BI-RADS-based class separation is frequently used (e.g, \cite{Geras2017,LiShenArxiv2018,dhungel2016automated,Neeraj17,Zhu_MGMIL_arXiv2016,Ayelet_MICCAIWS2016,Ayelet_MICCAIWS2017}) often because of the lack of pathological results in the dataset and the need to construct a large positive set. In Shen \etal \cite{LiShenArxiv2018}, the authors claimed that although the INbreast dataset includes pathology results, they use BI-RADS assessments for class labels, due to "lack of reliable pathological confirmation". In a similar way, they defined all images with BI-RADS 1 and 2 as negative and BI-RADS 4, 5 and 6 as positive.

Our second test bed used for our weakly supervised model, was composed of the INbreast (INB) publicly available FFDM dataset \cite{moreira2012inbreast}. This small dataset has 410 mammograms from 116 cases and was split into 100 positive (global BI-RADS 4,5,6) and 310 negative (global BI-RADS 1,2,3) mammograms. Note that in this case we included BI-RADS 3 to enable comparison with previous methods in literature. We conducted a random patients split on the INbreast images with 50\% for train and 50\% for test.

{\bf Implementation.} We implemented our model in the TensorFlow framework using the Adam optimizer for training, with a learning rate of $10^{-4}$, dropout of $0.5$, $l_2$-regularization and a batch-size of 256 images. This included all the regions from each image (on average approx. 200). We initialized the weights of the shared fully connected (FC) layer with a normal distribution \cite{glorot2010understanding}.
The weights of the FC layers in the branches were initialized with zero mean and $10^{-4}$ STD normal distribution.
For the number of selected regions we chose $k=10$ (other values were tested but yielded lower performance).
We set $\lambda_2=1$ in our semi-supervisded experiments, and we discuss other values in Sec. \ref{sec:exp_det}.
To enlarge and balance the training set, we used augmentations by adding rotations of $7 \times 45^{\circ}$, left-right and up-down flips and 6 image shifts.

{\bf Evaluation Procedure.}
Our evaluation on IMG dataset was based on 5 fold patient-wise cross-validation, where at each train and test iteration, all the images from the patient under test were strictly excluded from the training set. To this end we randomly split the dataset into 5 folds according to patient IDs, maintaining a similar distribution over breast composition and lesion types in the folds. All the performance values were based on the average over random split, 5-fold cross validation.

{\bf Compared Models.}
As our model outputs two probabilities per image ($p(y_c=1|x)$, Eq. \eqref{Eq:GlobalScores}), we can create 2D probabilities maps and conduct multi-class classification. However, to compare our results to previous methods and as an instance of a practical use case, we evaluated system performance on two binary classification tasks by joining two ``nearby" classes; namely, M with B or B with N. To this end we used $p(y_M=1|x)$ scoring for M vs. B $\cup$ N (M vs. BN) and $max\{p(y_M=1|x),p(y_B=1|x)\}$ scoring for M $\cup$ B vs. N (MB vs. N). For performance measures, in addition to AUROC, we also report two other practical measures as used in \cite{DreamChallenge}. The partial-AUC ratio (pAUCR) associated with the ratio of the area under the ROC curve in a high sensitivity range ([0.8,1]) represents the AUROC in a more relevant domain for clinicians.  In addition, we report the specificity extracted from the ROC curve at sensitivities of 0.85 and 0.90 that represent an average operation point (OP) for expert radiologists, as reported in \cite{BCSC2017}.

\subsection{Classification Results}
 We compare of our model's performance to several baselines.  We then discuss the impact of the fully labeled data engaged with our multi-task loss. 
 For evaluation, we present our results on the two binary classification tasks, M vs. BN and MB vs. N. In addition to the proposed Cls-Det-RS model, we implemented three baselines, 1) {\it Max-Region} \cite{choukroun2017mammogram} presenting a region classification only approach with max over regions, 2) the {\it DB-Baseline} presenting a dual-branch approach equivalent to Bilen \etal \cite{bilen2016weakly} and 3) the {\it Cls-Det} as our approach without region selection.

{\bf Weakly-supervised setup.} Table \ref{table:Results} presents performance for the two binary classification tasks. 
Considering purely the weakly labeled dataset, our method (Cls-Det-RS) outperformed the DB-Baseline and Max-region \cite{choukroun2017mammogram} on all measures and in both classification scenarios. 
The results of the model without the region selection (RS)  showed that in average, the addition of region-selection indeed improved performance.
We further conducted a breast level analysis by considering both views of the same breast. To this end we assigned the max probability between the views to the specific breast. The results exhibited similar performance to the single mammogram processing. 

Train and test on the small public data set of INB yielded AUROC of 0.73. Note that this result is without using an external {\bf fully labeled} data set in oppose to \cite{Lotter_MICCAI17,D_Ribli2018}. This result shows the performance of our model when trained on a very small data set. It is further comparable to AUROC 0.74, reported in \cite{Neeraj17} when trained on single MG, yet used fully supervised data. 

\begin{table*}
\begin{center}
\begin{tabular}{|l| c| c| c c|}
\hline
Method &  AUROC & pAUCR & \multicolumn{2}{c|}{Spec @ Sens} \\
&  	   &      & 0.85 & 0.90 \\
\hline
\multicolumn{5}{c}{\bf M vs. BN: Weakly-supervised methods} \\
\hline
DB-Baseline \cite{bilen2016weakly} & 0.709$\pm$ 0.020 & 0.251$\pm$0.05 & 0.37 &0.27 \\
Max-Region \cite{choukroun2017mammogram} & 0.699$\pm$ 0.047 & 0.235$\pm$0.10 & 0.36 &   0.24 \\
Cls-Det 	& 0.710$\pm$ 0.026 & \bf{0.280}$\pm$0.06 & \bf{0.42} &\bf{0.31} \\
Cls-Det-RS 	& \bf{0.728}$\pm$ 0.036 & 0.275$\pm$0.10 & 0.40 &   0.27 \\
\hline
\multicolumn{5}{c}{\bf MB vs. N: Weakly-supervised methods} \\
\hline
DB-Baseline \cite{bilen2016weakly} & 0.826$\pm$ 0.01 & 0.347$\pm$0.03 & 0.51 & 0.37 \\
Max-Region \cite{choukroun2017mammogram}& 0.817$\pm$ 0.02 & 0.323$\pm$0.07 & 0.48 & 0.35 \\
Cls-Det 	& 0.832$\pm$ 0.02 & 0.355$\pm$0.06 & 0.51 & 0.36 \\
Cls-Det-RS 	& \bf{0.841}$\pm$ 0.02 & \bf{0.367}$\pm$0.05 & \bf{0.55} & \bf{0.38} \\
\hline
\multicolumn{5}{c}{\bf M vs. BN: Semi-supervised methods} \\
\hline
 \hline
SS Cls-Det-RS .25	& 0.731$\pm$ 0.029 & 0.305$\pm$0.108 & 0.40 &   0.31 \\
SS-Baseline-RS .5	& 0.740$\pm$ 0.022 & 0.316$\pm$0.126 & 0.43 &   0.30 \\
SS Cls-Det-RS .50	& \bf{0.745$\pm$ 0.032} & 0.313$\pm$0.119 & \bf{0.46} &   \bf{0.33} \\
SS Cls-Det-RS .75	& \bf{0.745$\pm$ 0.026} & \bf{0.320$\pm$0.109} & 0.42 &   \bf{0.33} \\
\hline
\multicolumn{5}{c}{\bf M vs. BN: Fully-supervised (on M class) method} \\
\hline
SS Cls-Det-RS 1.0	& 0.751$\pm$ 0.026 & 0.316$\pm$0.078 & 0.47 &   0.32 \\

\hline
\end{tabular}
\end{center}
\caption{Binary classification performance compared to previous methods in weakly,  semi and full supervised setings.}
\label{table:Results}
\end{table*}

{\bf Semi-supervised setup.} Next we analyze the performance of our semi-supervised model. In order to reduce the demand for local annotations, we only considered local annotations for the malignant findings in our setting. We opted for the classification task of M vs. BN as commonly considered in previous works \cite{hwang2016self,old_DBLP:conf/miccai/ZhuLVX17,Neeraj17}. We further evaluated the impact of the ratio of the fully supervised train set as a measure of the cost effectiveness of the annotation workload. The results for our semi-supervised setting (Cls-Det-RS) are shown in Table \ref{table:Results}. The classification performance improved as more localized regions are used. This continued up to 100\% utilization of the local annotation (fully supervised).

\begin{figure*}[!ht]
\centering
{\includegraphics[width=1.7 cm, height=2.38 cm]{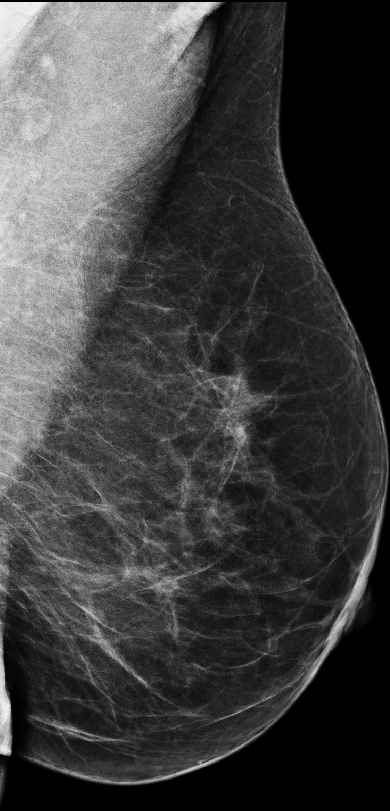}} 
{\includegraphics[width=1.7 cm, height=2.38 cm]{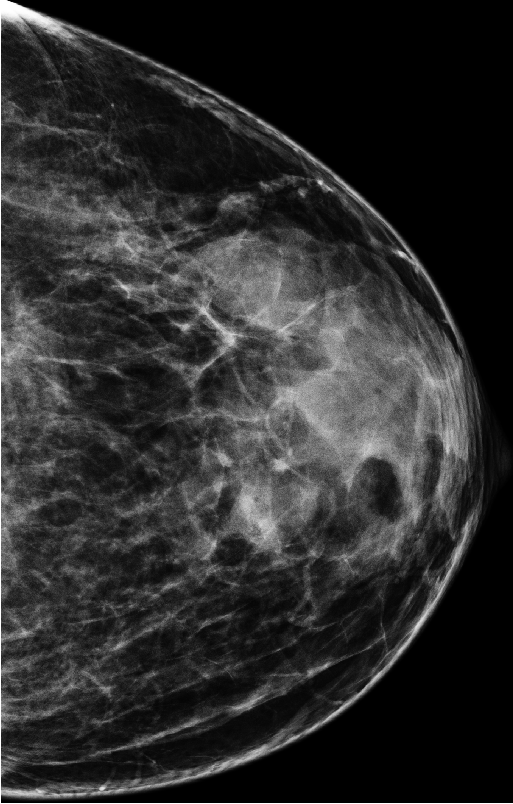}}
{\includegraphics[width=1.7 cm, height=2.38 cm]{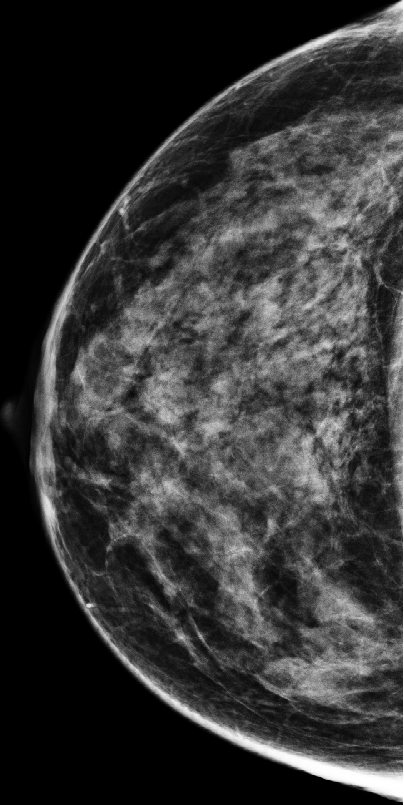}}
{\includegraphics[width=1.7 cm, height=2.38 cm]{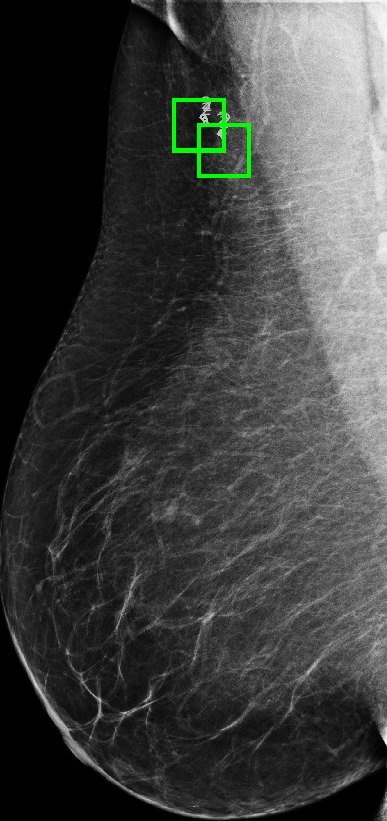}}
{\includegraphics[width=1.7 cm, height=2.38 cm]{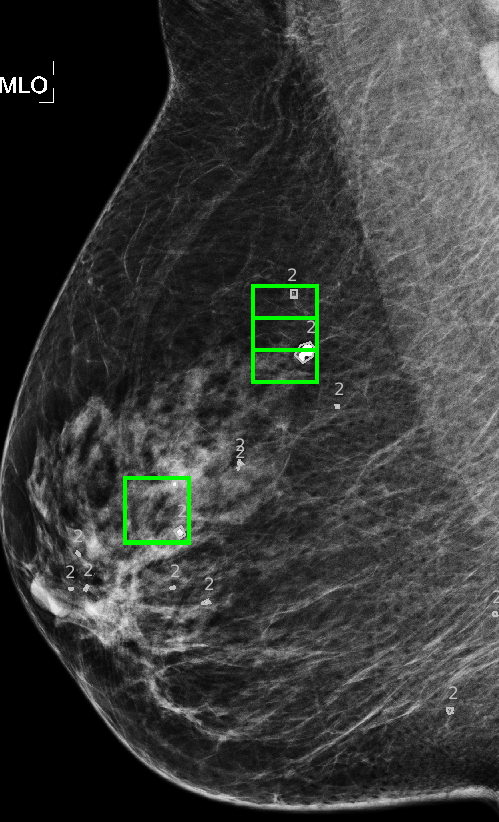}}
{\includegraphics[width=1.7 cm, height=2.38 cm]{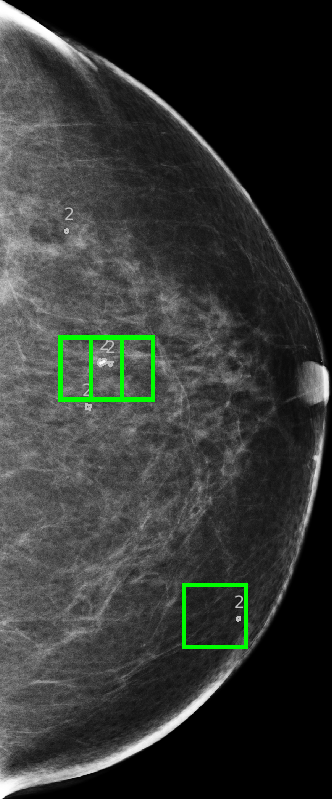}}\\
{\includegraphics[width=1.7 cm, height=2.38 cm]{detectionM1.png}}
{\includegraphics[width=1.7 cm, height=2.38 cm]{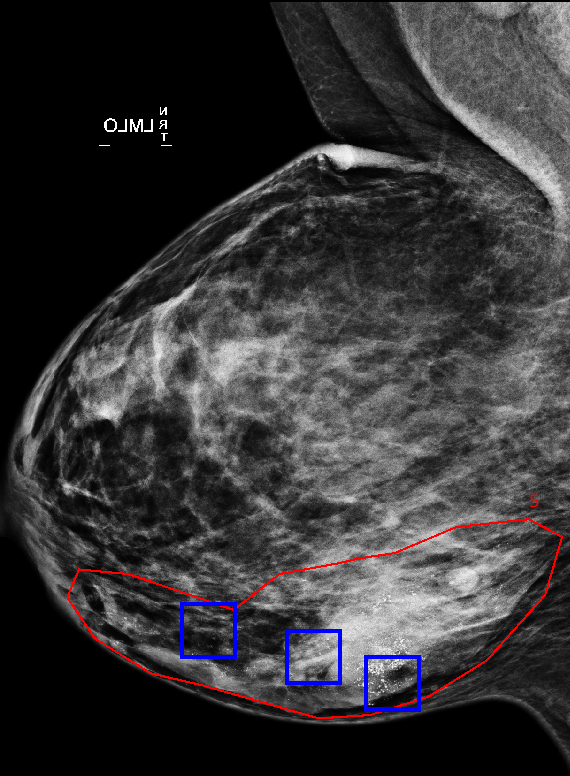}}
{\includegraphics[width=1.7 cm, height=2.38 cm]{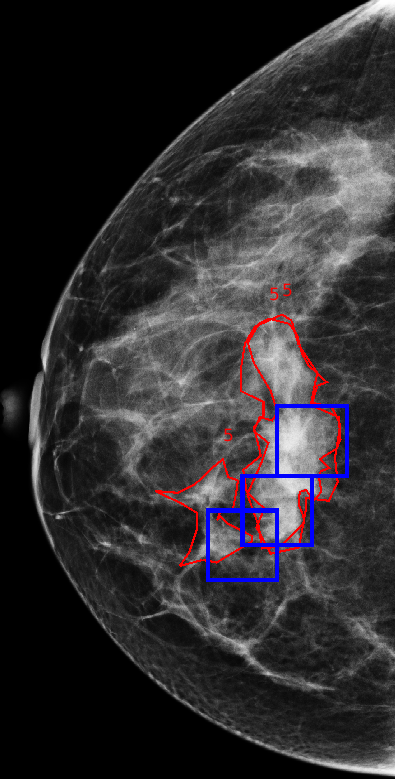}}
{\includegraphics[width=1.7 cm, height=2.38 cm]{detectionMB1.png}}
{\includegraphics[width=1.7 cm, height=2.38 cm]{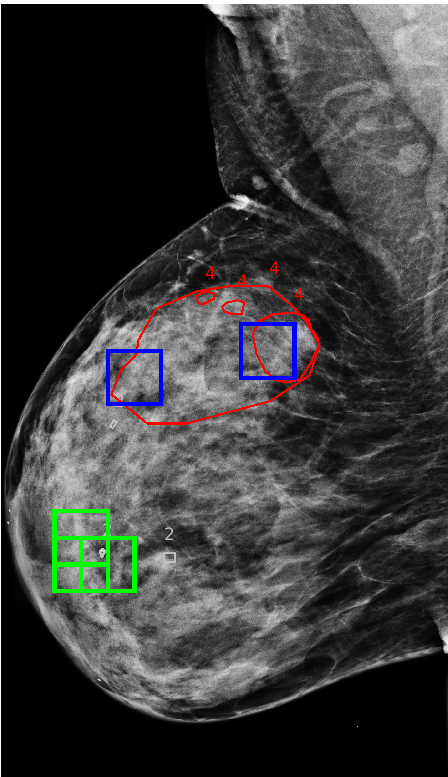}}
{\includegraphics[width=1.7 cm, height=2.38 cm]{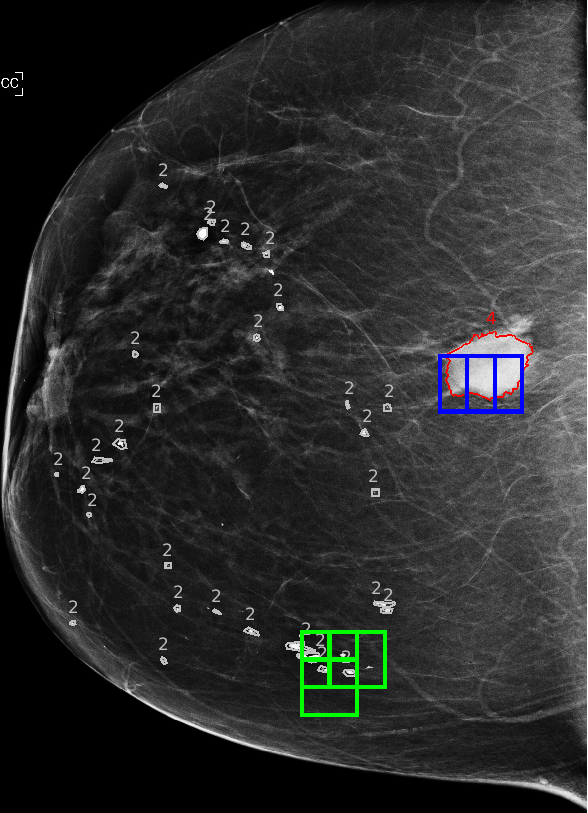}}
\caption{ Localization success in the weakly labeled setting. True malignant and benign lesions are annotated in red and gray respectively. Top 3 - M (blue) and B (green) regions are shown. Note the correlation between the radiologist's annotation and the model's predictions for each class. The top 3 images on the left are normal images without findings, where no bounding boxes were predicted. The 3 right hand images in the top row show cases of only benign findings. The lower 3 left hand images only have malignant findings, and the 3 right hand images have both malignant and benign findings. Note the agreement between the ground truth location and class of the finding with our predictions, without having any instance annotations in the training set. Best viewed in color.}
\label{fig:loc4fig}
\end{figure*}

\begin{figure*}[!ht]
\centering
\centering
\includegraphics[scale=0.295]{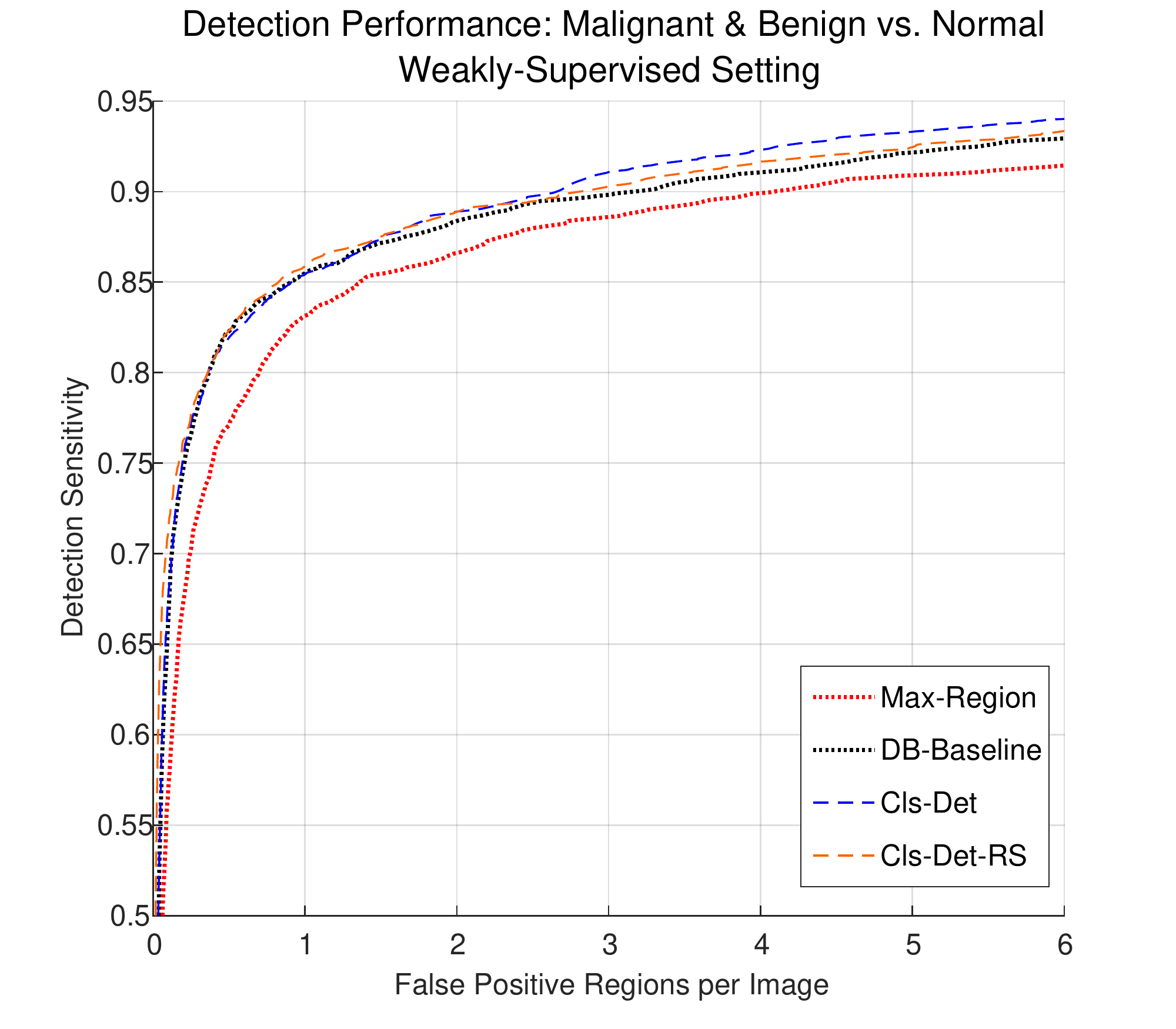}
\vspace{5mm}
\centering
\includegraphics[scale=0.295]{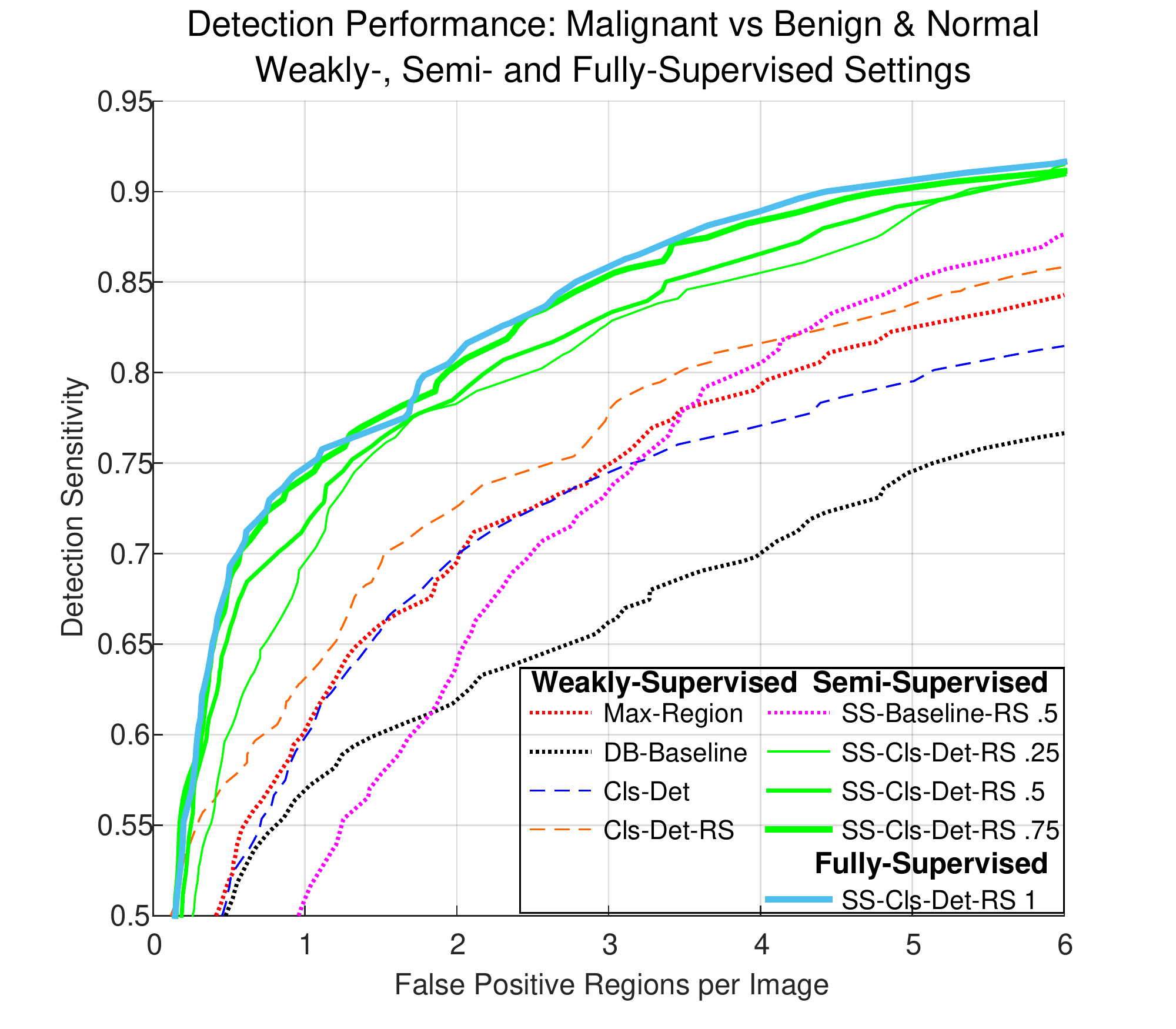}
\caption{FROC for detection performance at an operation point of 0.85 sensitivity in classification. Left: MB vs. N for the weakly-supervised setting. Right: M vs. BN, comparing weakly-, semi- and fully- supervised settings. Baselines: Max-region \cite{choukroun2017mammogram}, DB-Baseline\cite{bilen2016weakly}. SS-Baseline-RS .5: our semi-supervised approach with 50\% fully-supervised data when the fully-supervised objective is only on the classification branch. Weakly supervised proposed methods with and without region selection: Cls-Det, Cls-Det-RS. Proposed semi-supervised methods:  SS-Cls-Det-RS with various ratios of fully supervised data (indicated by green line with increasing width as a function of the fully-supervised data ratio). Fully-supervised method: SS-Cls-Det-RS 1. Best viewed in color.}
\label{fig:locRFFig}
\end{figure*}

\subsection{Detection Results} \label{sec:exp_det}
Although the train process begins without any labels on regions, the impact of each region can be scored after the training process by:
\begin{equation}
d^c(r_i) = p_{cls}(c|r_i) p^c_{det}(i|x), \quad c \in \{B,M\}, \quad i \in \{1,...,m\}
\label{eq:localMeasure}
\end{equation}
The top $k$ regions for each class (B/M) can now be visualized and compared to the radiologist's annotations as the source of malignancy or benign class of the image. 
Fig. \ref{fig:loc4fig} shows several examples with localization in the test set, overlaid with the radiologist's annotations (used only for validation).
As observed, the method is capable of separately highlighting multiple types of abnormalities such as benign and malignant lesions without having an instance level annotation.

We further evaluated our localization performance by a quantitative measure. Targeting the localization as the system's self-explanation tool, we used a less strict measure than the standard intersection over union (IoU) for correctness of our localization outcome. We follow the weak localization as intersection over the {\it minimum} area between the region and the lesion (IoM) as defined in Eq. \eqref{eq:IOM} (also used in \cite{Ben-AriISBI17}). This measure allows explanation of an outcome when a specified region contains a true type of lesion or vice-versa. Since our region size is relatively small and fixed, this setup will not allow over-sizing of the localization area (see examples in Fig. \ref{fig:loc4fig}). Unlike previous methods of \cite{hwang2016self,old_DBLP:conf/miccai/ZhuLVX17} we formally asses the accuracy of our localization results by Eq. \eqref{eq:localMeasure}.

For an image classified as $c$, we consider all the regions with $d^c(r_i)$ over a certain threshold. Correct localization per lesion is obtained if $IoM \geq 0.5$. We present the free-response receiver operating characteristic (FROC) localization accuracy for class $c \in \{M,B\}$ using $d^c(r_i) \geq Threshold$. The detection sensitivity in the FROC is the fraction of images in the True-Positive set with at least one correct localization. The results show that the region selection yielded the best performance with relatively low False positive per image (FPPI). 

{\bf Weakly-supervised setup}.
Fig. \ref{fig:locRFFig} shows the detection performance as FROC. Performance for MB vs. N is shown on the left. Although at low FPPI, DB-Baseline (dotted black curve) and our model (Cls-Det-RS, dashed orange curve) are comparable, at high detection sensitivity our model shows slightly improved performance. However, our model clearly
outperforms Max-Region \cite{choukroun2017mammogram} (dotted red curve).

Fig. \ref{fig:locRFFig} right plot depicts FROC curves for detection of the malignant lesions (BI-RADS 4 \& 5). In this set-up, we first compare our weakly supervised model to several baselines and then show the impact of our semi-supervised network with various ratios of fully labeled data. In particular, the detection performance in our weakly supervised model (dashed orange) is compared with the DB-Baseline \cite{bilen2016weakly} (dotted black) and the Max-Region method \cite{choukroun2017mammogram} (dotted red). In this scenario of detecting malignant lesions, the DB-Baseline shows poor results. Although the Max-Region shows improvement over DB-Baseline, our model clearly outperforms both. In addition, our model with region selection (Cls-Det-RS, dashed orange), outperforms our model without region selection (Cls-Det, dashed blue).

{\bf Semi-supervised setup}. 
The right plot in Fig. \ref{fig:locRFFig} shows that including local annotations in our semi-supervised model (SS-Cls-Det-RS) improves detection. The green lines indicate results when using different ratios of fully labeled data (wider curves indicate higher fully labeled ratio in training). The wide cyan line stands for full supervision on the {\bf M class}. The detection sensitivity further improved when more locally annotated mammograms were used. However, the influence of local annotations plateaus approaching the 75\% ratio (SS-Cls-Det-RS 0.75), presenting similar performance to the fully-supervised method (SS-Cls-Det-RS 1).
The performance drop in M vs. BN compared to MB vs. N (right vs. left plot in Fig. \ref{fig:locRFFig}), indicates the model's difficulty in distinguishing between benign and malignant lesions, as often is the case with radiologists.

\begin{figure*}[!ht]
\centering
\includegraphics[scale=0.295]{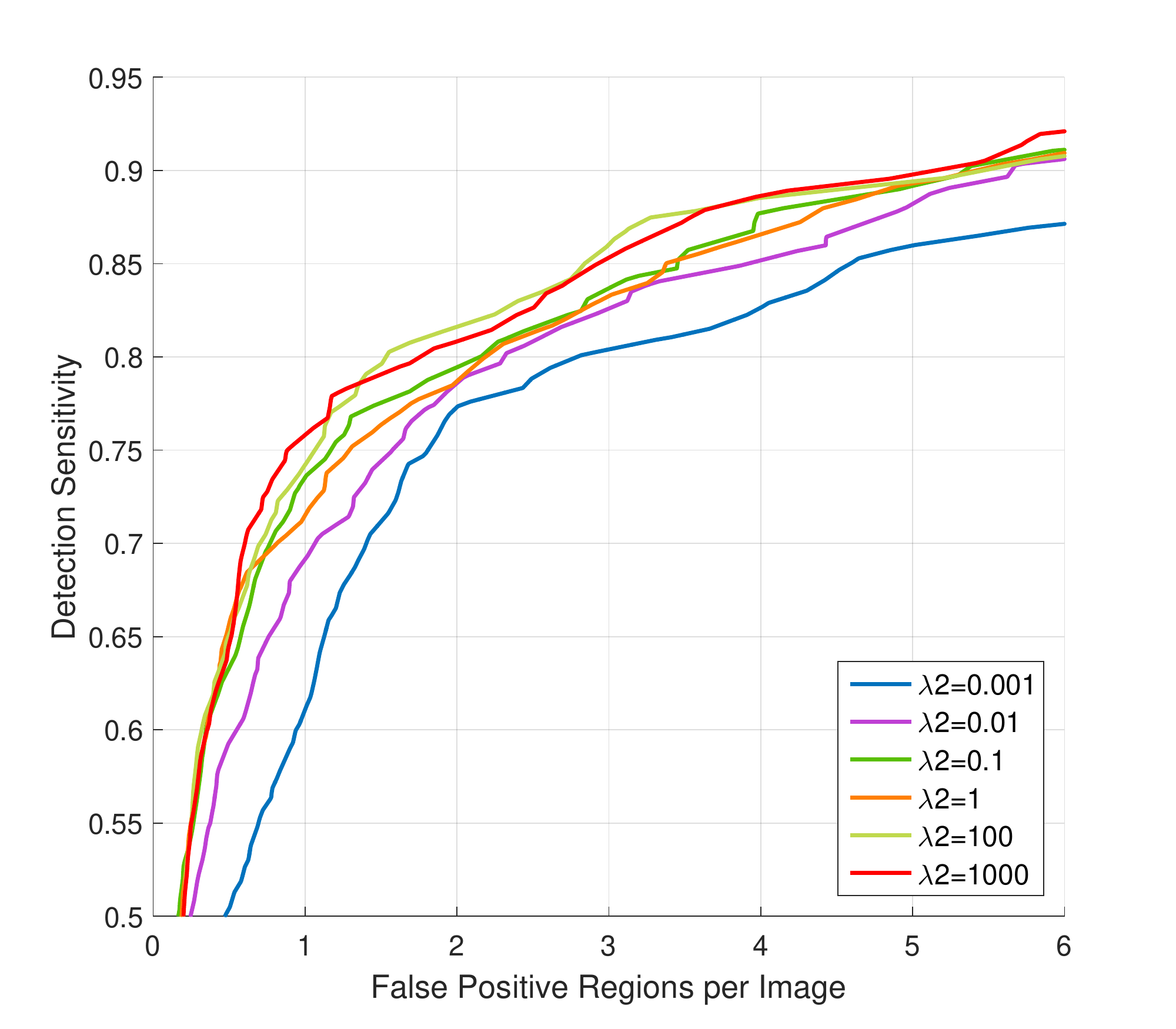}
\caption{FROC of M vs. BN detection performance at an operation point of 0.85 sensitivity in classification for several values of $\lambda_2$.}
\label{fig:locRFFigLambda}
\end{figure*}

{\bf Setting the value of $\lambda_2$.} The parameter  $\lambda_2$ controls the balancing between the fully supervised images and the weakly supervised images (that contain both benign and malignant images). We found that the classification and localization result are insensitive 
to the values of $\lambda_2$ in a range of roughly $0.1\!-\!1$. Increasing $\lambda_2$ yields lower performance in the benign class without improving the classification performance of the malignant class and with slight improvement in the localization of the malignant class (e.g. for $\lambda_2=5$, we get 0.72 MB vs. N AUC, and for $\lambda_2=10$, this decreases to 0.65 AUC). As decreasing $\lambda_2$, we get a slightly worse classification performance of M vs. BN and worse performance in the localization.  Fig. \ref{fig:locRFFigLambda} shows FROC of  our semi-supervised approach, SS-Cls-Det-RS, with 50\% fully-supervised data for various values of $\lambda_2$.

{\bf The impact of loss on the detection branch.}
To this end, we ran our model with loss solely on the classification branch (similar to \cite{DBLP:journals/corr/YanLPLZ17}). We trained our model with 50\% fully labeled data, without the detection loss in Eq. \eqref{eq:fullySupervisedObj} (setting $\mathcal{L}^{\mathcal{F}}_{Det}(\theta)=0$). The resulting FROC (SS-Baseline-RS .5 - dotted pink curve) appears in Fig. \ref{fig:locRFFig}-right. Comparison to our model (SS-Cls-Det-RS .5 - green) indicates a significant drop of FROC in this baseline, and points to the contribution of our novel detection loss.

\subsection{Further Analysis on Classification Results}
In this subsection, we presents a visual analysis of the classification results of our weakly supervised method, and multi-class classification results.

\begin{figure*}[!ht]
        \centering
        \includegraphics[scale=0.295]{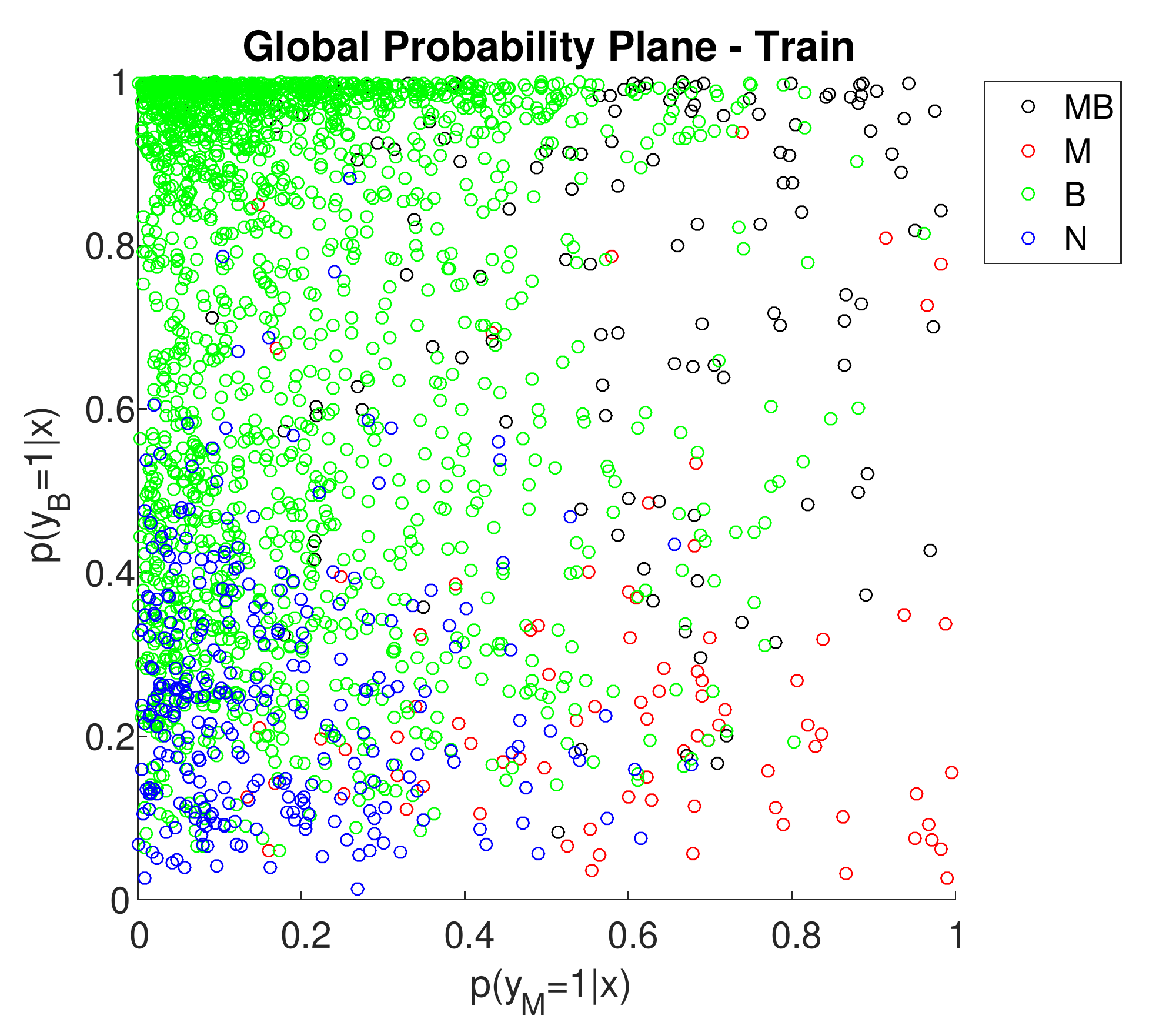}
        \includegraphics[scale=0.295]{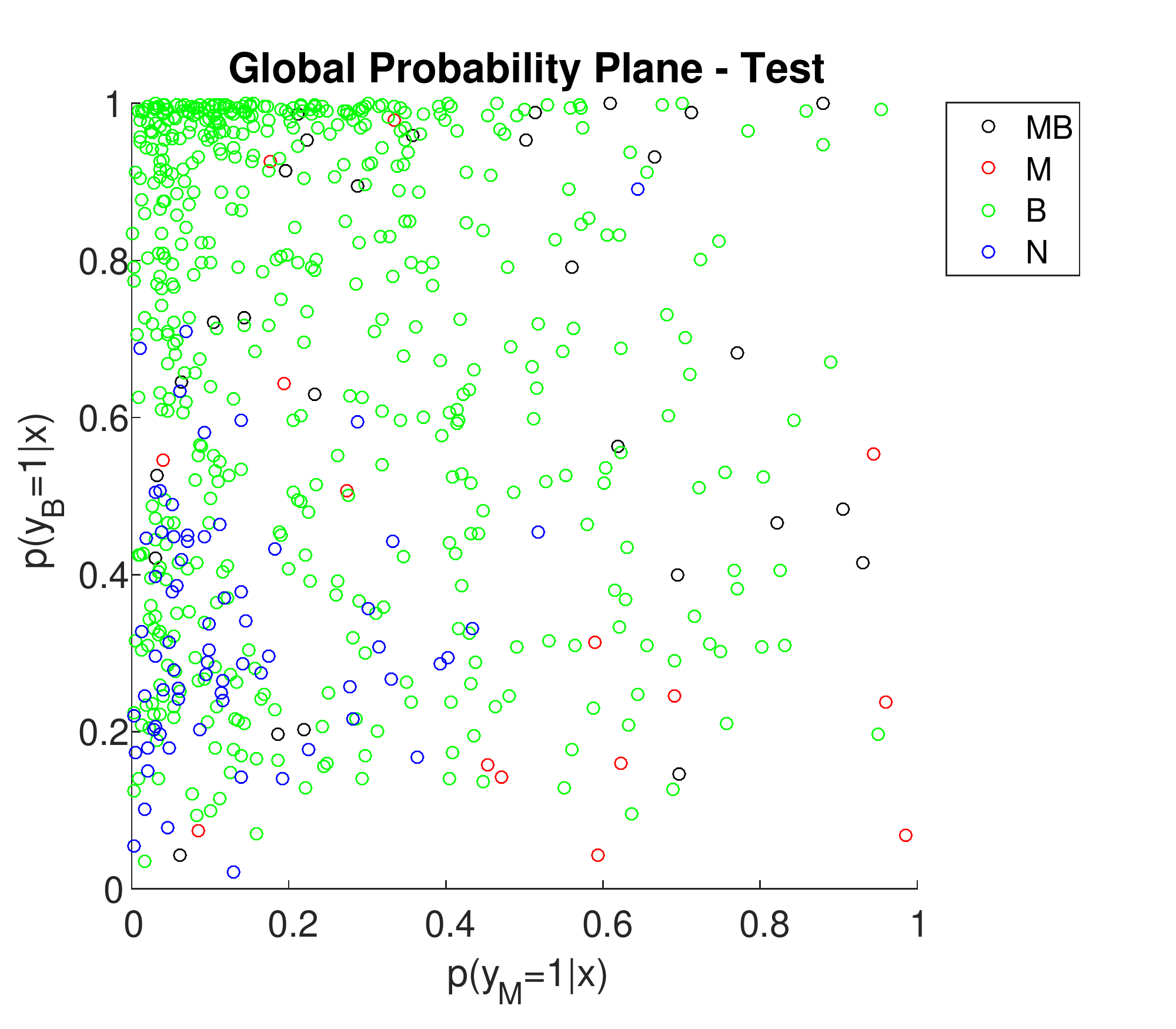}
\caption{Global probabilities plane for the train and test set. The figure shows the probability of at least one malignant or benign finding appearing in the mammogram. The samples are color coded to depict prediction accuracy, N-Normal, B-Benign, M-Malignant and MB-Benign and Malignant findings present in the mammogram. Best viewed in color.}
\label{fig:GlobalProbPlane}
\end{figure*}

Using our multi-label probability output we plot each sample in a probability plane representing the global prediction results of the images. In this plane, each image is a $2D$ point with coordinates as $p(y_M=1|x)$ and $p(y_B=1|x)$ probabilities. Fig. \ref{fig:GlobalProbPlane} shows the global probability plane on a train and test set color coded by the true class. Blue normal (N) images (without any finding) are mostly located near the origin, with low $p(y_M=1|x)$ and $p(y_B=1|x)$ showing approximately zero probabilities for malignancy and benign. Green represents images with only benign findings (B). Those are likely concentrated around (0,1) with low $p(y_M=1|x)$ and high $p(y_B=1|x)$. Red points represent malignant images without benign findings. Those are emerged at the right side of the plot with high $p(y_M=1|x)$ and low $p(y_B=1|x)$. Finally, black points, representing malignant mammograms that also include benign findings, are more likely located in the top-right corner with high $p(y_M=1|x)$ and high $p(y_B=1|x)$.

\section{Conclusion}
In this work, we proposed a method for multi-class classification of mammograms and detection of abnormalities in weakly and semi-supervised settings. We addressed the problem of fusion between weak labels and local annotations in the dataset via a novel objective function. As local annotations are prohibitively expensive in the medical domain, our semi-supervised approach allows reaching nearly fully labeled data performance with a fraction of local annotations. The new model relies mainly on weakly labeled data and therefore can run without any local annotations is the dataset. 

We demonstrate our method on a large dataset, and compare our approach with various measures, to several baselines and as well as direct comparison to a previously published method.  The results show improvement in AUROC, with a significant performance boost in partial AUC and a practical operation point. Locally annotating only 5\% of the data yielded a 10\% increase in specificity (at 0.85 sensitivity) that is estimated to lead to to yearly 3.6 million fewer false positives in screening mammography \cite{BCSC2017}. Our method can learn solely from image-level labels, and utilize possibly existing local annotations as bounding boxes around lesions.

A major feature of our system is the localization of the image  level decision.
This makes system decision interpretable to physicians who obtain the automatic decison. We evaluated our localization performance {\it quantitatively}, in {\it full resolution}. The results compete favorably with a previous weakly supervised method and significantly improve in our semi-supervised approach. In the era of Big Data, the combination of large weakly labeled data sets with partially local annotations can provide a cost-effective solution for future decision support systems in medical imaging.

Possible applications suggest second reader in screening mammography and other imaging domains. System explanation based on lesion localization and category should encourage trust among radiologists and is necessary in cases where a quick over-rule is needed if the system decision was found to be wrong.

Our method was evaluated, based on BI-RADS assessment by radiologists.  We opted for this setting in order to have a large dataset of approximately  3K mammograms, as pathologies were not available for all of our high BI-RADS exams. BI-RADS 4 and 5 have positive predictive values of approx. 35\% and over 95\% respectively and are particularly rare in the population. There are several recent works trained and tested on large FFDM mammogram datasets with pathologies such as \cite{Lotter_MICCAI17} and \cite{D_Ribli2018} which used the DREAM Challenge dataset, or \cite{Geras2019}. Unfortunately, these datasets are not publicly available and cannot be used by other researchers for benchmarking. We believe that our scenario based on BI-RADS assessments can provide a valid platform for comparison between different methods and baselines. We tested our method and all compared methods on the same data setting to allow for fair comparison.

Our method was limited to analyzing each view separately, without bilateral breast comparison as conducted by radiologists. We intend to use this additional information in our future work to extract correlations between image views and dissimilarities between breast sides.

Combining the proposed approach with end-to-end training of the backbone network is applicable with larger datasets.  End to end training as well as using multiple scale and aspect ratio regions constitute interesting future research directions that are beyond the scope of this work.

% \section*{References}
\bibliographystyle{elsarticle-num}
\bibliography{ref}{}

\end{document}